\documentclass[10pt,twocolumn,letterpaper]{article}

\usepackage{iccv}
\usepackage{times}
\usepackage{epsfig}
\usepackage{graphicx}
\usepackage{amsmath}
\usepackage{amssymb}

\usepackage{booktabs}
\usepackage{url}
\usepackage{enumitem}
\usepackage{tikz}
\usepackage{comment}
\usepackage{color}
\usepackage{textcomp}  
\usepackage{amsfonts}
\usepackage{multirow}
\usepackage{algorithm}
\usepackage{algorithmic}
\usepackage{float}
\usepackage{tabularx}
\usepackage{array}
\usepackage[misc]{ifsym}
\usepackage{caption}
\usepackage{lipsum}
\usepackage{wrapfig}
\usepackage{makecell} 

\usepackage{amsmath,amsfonts}
\def\bm#1{\mathbf{#1}}

\def\eqref#1{equation~\ref{#1}}

\def\1{\bm{1}}

\def\vx{{\bm{x}}}

\def\mF{{\bm{F}}}
\def\mG{{\bm{G}}}

\def\mM{{\bm{M}}}

\def\mR{{\bm{R}}}

\def\mX{{\bm{X}}}

\DeclareMathAlphabet{\mathsfit}{\encodingdefault}{\sfdefault}{m}{sl}
\SetMathAlphabet{\mathsfit}{bold}{\encodingdefault}{\sfdefault}{bx}{n}

\usepackage{xcolor}
\usepackage{soul}
\definecolor{my_blue}{HTML}{2b90d9}
\definecolor{my_red}{HTML}{ff5f2e}
\definecolor{my_yellow}{HTML}{FFF2CC}

\usepackage[pagebackref=true,breaklinks=true,letterpaper=true,colorlinks,bookmarks=false]{hyperref}

\iccvfinalcopy 

\def\iccvPaperID{1506} 

\ificcvfinal\fi

\begin{document}

\title{TaskExpert: Dynamically Assembling  Multi-Task Representations \\ with Memorial Mixture-of-Experts}

\author{Hanrong Ye and Dan Xu\textsuperscript{\Letter}\\
Department of Computer Science and Engineering, HKUST\\
Clear Water Bay, Kowloon, Hong Kong\\
{\tt\small \{hyeae, danxu\}@cse.ust.hk}
}

\maketitle
\ificcvfinal\thispagestyle{empty}\fi

\begin{abstract}
Learning discriminative task-specific features simultaneously for multiple distinct tasks is a fundamental problem in multi-task learning. Recent state-of-the-art models consider directly decoding task-specific features from one shared task-generic feature (e.g., feature from a backbone layer), and utilize carefully designed
decoders to produce multi-task features. However, as the input feature is fully shared and each task decoder also shares decoding parameters for different input samples, it leads to a static feature decoding process, producing less discriminative task-specific representations.
To tackle this limitation, we propose TaskExpert, a novel multi-task mixture-of-experts model that enables learning multiple representative task-generic feature spaces and decoding task-specific features in a dynamic manner. Specifically, TaskExpert introduces a set of expert networks to decompose the backbone feature into several representative task-generic features.
Then, the task-specific features are decoded by using dynamic task-specific gating networks operating on the decomposed task-generic features.
Furthermore, to establish long-range modeling of the task-specific representations from different layers of TaskExpert, we design a
multi-task feature memory that updates at each layer and acts as an additional feature expert for dynamic task-specific feature decoding.
Extensive experiments demonstrate that our TaskExpert clearly outperforms previous best-performing methods on \textbf{all 9 metrics} of two competitive multi-task learning benchmarks for visual scene understanding (i.e., PASCAL-Context and NYUD-v2).
{Codes and models will be made publicly available \href{https://github.com/prismformore/Multi-Task-Transformer}{here}.}
\end{abstract}

\section{Introduction}
\begin{figure}[t]
    \centering   \includegraphics[width=0.98\linewidth]{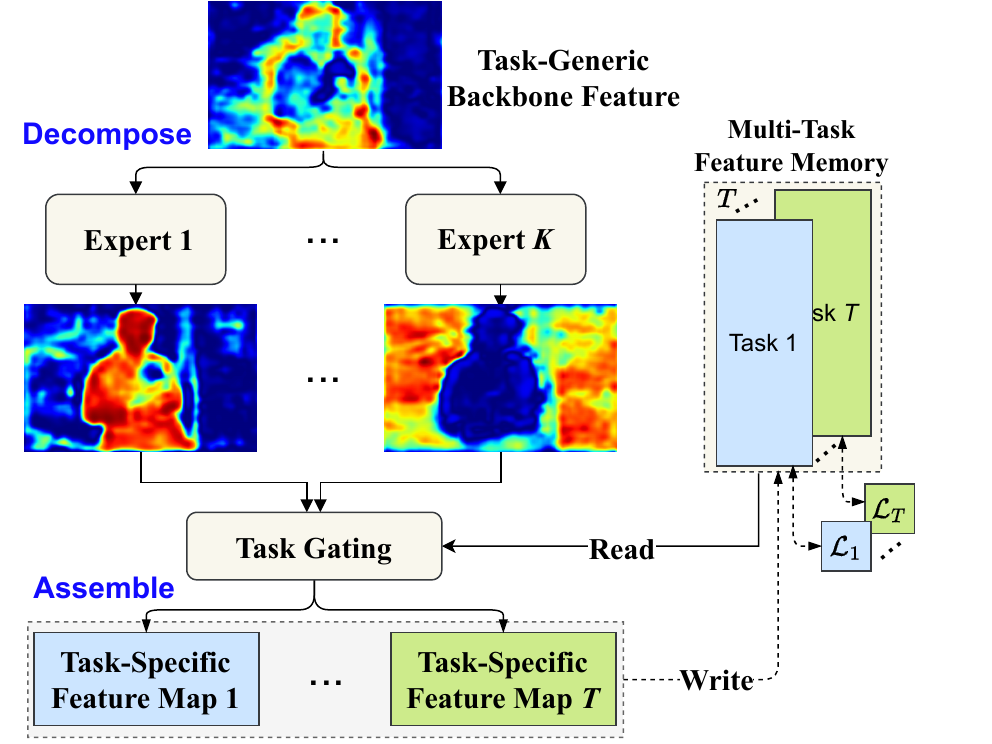}
    \vspace{-5pt}
    \caption{
    Motivative illustration of TaskExpert for dynamical decoding discriminative task-specific representations. A set of $K$ expert networks decompose the task-generic backbone feature into $K$ representative feature spaces.
    Then, the feature activations of an input from the multiple expert networks are dynamically assembled by the Task Gating network, to produce task-specific features for the $T$ tasks.
    TaskExpert further performs long-range modeling of task-specific representations by designing a multi-task feature memory. It is updated at the current layer based on the assembled task-specific representations, and is used as an additional feature expert in the next layer.
    }
    \label{fig:concept}
    \vspace{-18pt}
\end{figure}

With the rapid development of deep learning models, many computer vision tasks can be effectively handled by deep networks, which strongly motivates researchers to develop unified multi-task networks to perform joint learning and inference on multiple distinct tasks~\cite{mti,atrc,zamir2020robust,kokkinos2017ubernet}.
Multi-Task Learning (MTL) can avoid repetitive training of single-task models
and is able to simultaneously generate predictions for different tasks with only one-time inference. On the other hand, as different computer vision tasks usually share a basic common understanding of input images,
learning multiple tasks concurrently can help improve the representation power of each task and boost the overall performance of multi-task models~\cite{astmt,kendall2018multi}.

\par Learning discriminative task-specific features is a fundamental problem for MTL models. To achieve this objective, recent state-of-the-art works~\cite{invpt2022,atrc,padnet,mti} consider a decoder-focused paradigm, in which a shared backbone model with pretrained parameters is applied to learn generic image representations, and then separate task decoders are carefully designed to generate task-specific features. Although this paradigm is straightforward to achieve task-specific feature decoding, and produces promising multi-task prediction results, these models have two aspects of potential limitations: (i)~The input task-generic feature from a backbone layer is fully shared for the different task decoders, which requires a careful design of the decoders~(\textit{e.g.}~capacity and structure) for the different tasks to achieve task-discriminative decoding. (ii)~The parameters of each task decoder are typically shared for different input samples, resulting in a static decoding of task-specific features, while a sample-dependant decoder is highly beneficial for learning more discriminative task features, as the input samples can be diverse, and sample-related context information is particularly important for multi-task learning and inference. All these aspects hamper the previous models from learning effective task-specific features and predictions.

\par To tackle the above-mentioned issues, this paper proposes a novel multi-task mixture-of-experts framework, coined as ``TaskExpert'', which dynamically performs the task-feature decoding for different inputs and tasks as depicted in Figure~\ref{fig:concept}.
Instead of using one shared backbone feature to decode different task features, the proposed TaskExpert introduces a set of task-generic expert networks that learn to \textbf{\textit{decompose}} the backbone feature into a set of representative task-generic features.~The feature decomposition enables the decoders to interact with finer-granularity task-generic feature spaces to produce more discriminative task features, as each expert can be responsible for modeling one representative feature space from the training data. Then, we dynamically \textbf{\textit{assemble}} feature activations from different experts to produce task-specific representations, based on sample-dependent and task-specific gating scores predicted from a designed gating network.
On the other hand,
visual understanding tasks greatly benefit from representations of different network levels~\cite{lin2017feature}. Thus, we further devise a ``multi-task feature memory'' to aggregate the dynamically decoded task-specific features produced at different network layers, and propose a ``Memorial Mixture-of-Experts (\textbf{MMoE})'' module. For each task, MMoE utilizes its corresponding task feature from the multi-task feature memory as an additional feature expert, to improve task-specific feature decoding.
In this way, MMoE enables long-range dynamic modeling of task-specific representation for each task throughout the entire backbone.

\par In summary, the main contribution is three-fold:
\begin{itemize}
\vspace{-5pt}
\item We propose a multi-task mixture-of-experts model that allows effective decomposition of a task-generic feature from any backbone network layer and enables dynamically assembling discriminative multi-task representations by a task gating mechanism.
\item
We design a novel Memorial Mixture-of-Experts~(MMoE) module by further introducing a multi-task feature memory, which can effectively interact~(read and write) with the mixture of experts at different network layers to achieve long-range dynamic decoding of task-specific features by utilizing different levels of network representations.
\item The proposed TaskExpert is extensively validated on two challenging multi-task visual scene understanding benchmarks (\textit{i.e.}~Pascal-Context and NYUD-v2). The results clearly show the effectiveness of TaskExpert for dynamic multi-task representation learning, and also establish new state-of-the-art performances {on all 9 metrics} on the benchmarks, regarding the comparison with previous best-performing methods using both transformer- and CNN-based backbones.
\end{itemize}

\section{Related Works}
\par\noindent\textbf{Multi-Task Learning in Computer Vision}
In the field of multi-task learning (MTL) for computer vision problems~\cite{zhang2021survey,mtlsurvey,zamir2018taskonomy,kanakis2023composite,bachmann2022multimae,hoyer2021three,liu2019MTAN}, some research works explore effective solutions through a perspective of model optimization and improve the optimization process of MTL via different multi-task loss designs~\cite{zamir2020robust,kendall2018multi,liu2021conflict,contrastiveMTL2023,li2022Learning} and gradient manipulations~\cite{gradientsign,gradnorm,wang2020gradient,yu2020gradient}. These optimization strategies have also been demonstrated beneficial for alleviating the issue of task competition in multi-task learning~\cite{kendall2018multi}.
From another perspective, several existing works also investigate designing better multi-task model architectures.~These related works can be roughly divided into two groups:~encoder-focused and decoder-focused methods.
The encoder-focused methods~\cite{crossstitch,nddr,zhang2021automtl,gao2020mtl} target designing multi-task encoders, while the decoder-focused methods typically share the encoder~(\ie the backbone) for different tasks and design sophisticated decoders for task-specific feature learning and prediction~\cite{mti,atrc,papnet,padnet,zhang2021transfer,psd}.
A recent state-of-the-art decoder-focused method~(\textit{i.e.}~InvPT~\cite{invpt2022}) performs multi-task learning on powerful transformer architectures~\cite{transformer}, and establishes a very strong baseline for multi-task visual scene understanding. It can effectively learn spatial and cross-task relationships globally.
{To adapt multi-task models to different tasks, researchers also propose to search for task-specific network structures for encoders~\cite{vandenhende2019branched,lu2017fully,guo2020learning} and decoders~\cite{atrc}, or utilize task-specific learnable prompts to effectively capture task-specific information~\cite{taskprompter2023}.}~Despite promising performances of these works, they are still limited by static decoder network designs since their decoders cannot dynamically generate task-specific representations based on different inputs and tasks. In this paper, we propose TaskExpert, a novel and effective mixture-of-experts framework, which can achieve dynamic multi-task representation learning via feature decomposition and assembling through the entire backbone network.

\par\noindent\textbf{Mixture-of-Experts~Models}~With~statistical~motivation~\cite{jordan1994hierarchical}, Mixture-of-Experts (MoE) models are originally designed to control the dynamics of neural networks automatically~\cite{jacobs1993learning,jacobs1991adaptive}. MoE learns a series of expert networks and a gating network. The outputs of expert networks are weighted by gating scores (or called ``gate values'') generated by a gating network before the weighting operation. In more recent works, some researchers use the gating scores as a criterion to sparsely select only one or a few experts. The sparse activation of experts enables a significant reduction of the computational cost when training large-scale models~\cite{eigen2013learning,shazeer2017outrageously,fedus2021switch,lepikhin2020gshard}.
Related to our work,~\cite{ma2018multigate} designs a multi-gate MoE to ensemble expert networks for different census analysis tasks with different gating networks, and~\cite{m3vit,chen2022mod} propose sparse task-conditioned MoE networks for efficient multi-task learning, which generate predictions for \emph{only one} task within a single forward network inference. In contrast to these works, our Memorial Mixture-of-Experts~(MMoE) is a plug-and-play decoder module, which is designed for dynamically assembling discriminative task-specific features. {Our TaskExpert can simultaneously generate multi-task predictions for all tasks in one single forward inference, leading to significantly higher multi-task training efficiency and enabling cross-task interaction learning. Moreover, MMoE models task-specific features throughout different network layers via the novel design of multi-task feature memory and help generate more discriminative task-specific features.}

\begin{figure}[t]
    \centering
    \includegraphics[width=.98\linewidth]{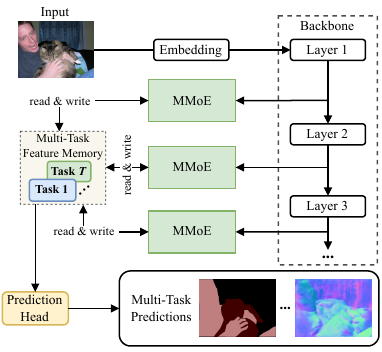}
    \vspace{-5pt}
    \caption{Overall architecture of the proposed TaskExpert model. Memorial Mixture-of-Experts (MMoE) modules are embedded at different backbone layers to dynamically decode task-specific feature for each task at the corresponding layer. The MMoE also interacts (reads and writes) with a designed multi-task feature memory to have a long-range dynamic multi-task feature decoding benefiting from different levels of network features. The reads utilize the memory as an additional feature expert in MMoE, and the writes update the memory at the layer. After being updated by the last network layer, the multi-task feature memory is used for the final prediction of each task.}
    \label{fig:architecture}
    \vspace{-15pt}
\end{figure}

\section{TaskExpert}
\subsection{Framework Overview}
An overall illustration of our TaskExpert framework is shown in Figure~\ref{fig:architecture}. The input image is first projected to a token sequence and then fed into a series of classic visual transformer layers. Our framework can also be flexibly used with a CNN backbone. The proposed Memorial Mixture-of-Experts~(MMoE) module is embedded at different network layers, and each module takes the task-generic backbone feature of that layer as input. It dynamically assembles and outputs discriminative task-specific representations at the corresponding layer.
The MMoE further interacts~(reads and writes operations) with the designed multi-task feature memory,  to perform a long-range dynamic multi-task feature decoding, which benefits from different levels of network features. The reads utilize the feature memory as an additional feature expert in the MMoE module, and the writes update the memory at the layer.
Finally, after being updated by the outputs of all MMoE modules at different network layers, the task-specific features in multi-task feature memory are then used for generating the multi-task predictions with simple prediction heads. We elaborate on more details of our model in the following several sections.

\subsection{Multi-Task Backbone Structure}
\label{sec:model_backbone}
We consider a widely-used vision transformer (ViT) model~\cite{vit} with $L$ layers as the backbone of our multi-task MoE model.
The input image is first divided into $H\times W$ patches and processed by the patch embedding and positional embedding modules. Then, we obtain a sequence of patch tokens $\mX^0 = \{\vx_1^0, \vx_2^0,...,\vx_N^0\}$, where $N=H\times W$ and the number of channels is $C$. The patch tokens contain generic visual representations of the input image. We denote the $l$-th transformer layer as a mapping function $f^l$, and the updating of image patch tokens at different transformer layers can be formulated as $\mX^l = f^l(\mX^{l-1})$, where $\mX^l \in \mathbb{R}^{N \times C}$ denotes the output patch tokens of the transformer at the $l$-th layer.

\begin{figure*}[t]
    \centering
    \vspace{-20pt}
\includegraphics[width=1\linewidth]{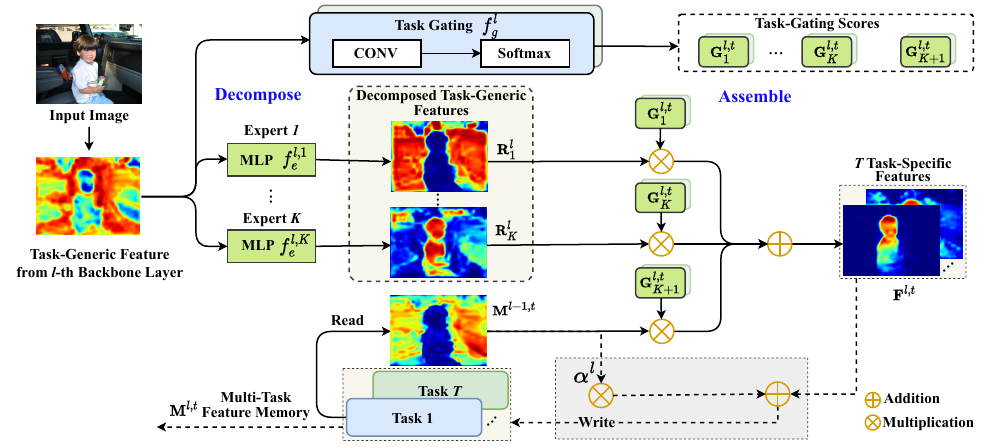}
    \vspace{-18pt}
    \caption{Illustration of the proposed Memorial Mixture-of-Experts (MMoE) module with the multi-task feature memory.
    MMoE accepts a task-generic feature from a backbone layer as input, and dynamically generates task-specific features for different tasks in a sample- and task-dependent manner.
    Specifically, MMoE decomposes the backbone feature to a set of $K$ representative task-generic features ($\mathbf{R}^l_1,...,\mathbf{R}^l_K$) via the $K$ expert networks. Then these features are used to assemble a set of $T$ task-specific features based on dynamically generated task-gating scores from the task gating networks.
    To establish long-range modeling of task-specific representations crossing different network layers, the multi-task feature memory is designed to interact with the expert networks. The multi-task feature memory serves as an additional feature expert to decode the task-specific representations, and the decoded representations are also further used to update the multi-task feature memory.
    }
    \label{fig:mmoe}
    \vspace{-10pt}
\end{figure*}
\subsection{Memorial Mixture-of-Experts~(MMoE)}
\label{sec:mmoe}
We present the details of several core parts of the proposed memorial mixture-of-experts (MMoE) module, including backbone generic-feature decomposition with a set of network experts, context-aware task-specific gating, the multi-task feature memory to have long-range modeling on task-specific features through interactions~(reads and writes) with the mixture of experts, dynamic task-specific feature assembling, and multi-task prediction heads.

\begin{figure*}[!t]
    \centering
    \vspace{-15pt}
     \includegraphics[width=1.\linewidth]{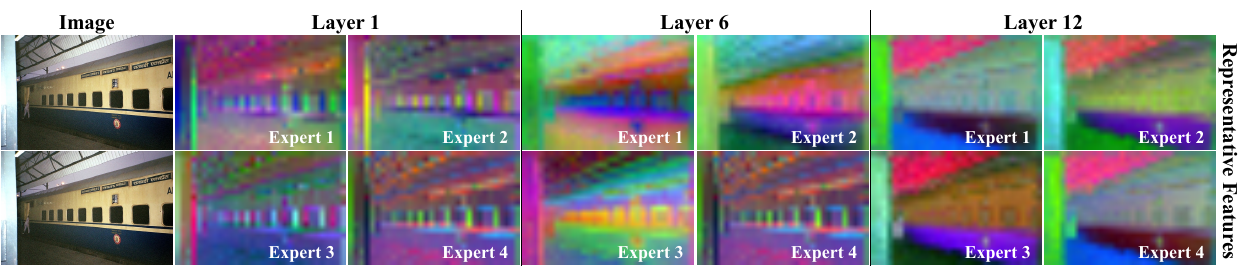}
     \vspace{-20pt}
     \caption{{PCA visualization of the representative features by task-generic experts across different layers.  In the deeper layers, the features become increasingly responsive to high-level semantic areas. Conversely, features at more shallow layers show a heightened sensitivity to low-level patterns.}}
     \label{fig:vis_rep_features_layers}
\vspace{-10pt}
\end{figure*}

\subsubsection{Backbone Feature Decomposition with Experts}
As shown in Figure~\ref{fig:mmoe}, to enable learning discriminative task-specific representations from the task-generic backbone feature, we first define a set of expert networks that decompose the backbone feature into several representative task-generic features in sub-feature spaces, representing distinct patterns of the backbone features, as can be observed in Figure~\ref{fig:vis_path_rep}. It allows more fine-grained decoding of task-specific features.
Specifically, at the $l$-th backbone layer, let us assume that we use $K$ expert networks $\{f_e^{l,i}(\cdot): \mathbb{R}^{N\times C} \rightarrow \mathbb{R}^{N\times C}, i\in [1,K]\}$ to learn distinct task-generic feature projections, where superscript $l$ denotes the layer index and $i$ is the expert index. Each expert network employs the same Linear-BatchNorm-RELU-Linear structure but with different parameters. Through each expert network, we obtain a set of representative features~$\{\mR_i^l \in \mathbb{R}^{N\times C}, i \in [1,K]\}$ by:
\begin{equation}
    \mR_i^l = f_e^{l,i}(\mX^{l}),
\end{equation}
where $\mathbf{X}^l$ is the $l$-th layer feature of the backbone.

\subsubsection{Context-Aware Task-Specific Gating}
After we have obtained a set of representative task-generic features decomposed from a backbone feature for all the tasks, we aim at decoding discriminative task-specific features for different tasks by automatically assembling these representative features. Thus, we propose to adaptively control the contribution of each expert for each patch token by designing a task-specific gating network.

\par For each token in the representative task-generic features, we compute a task-gating score to control its contribution to generating the task-specific features.
The task-gating scores of each task are dynamically generated by a task-specific gating network, using the features $\mathbf{X}^l$ from the $l$-th network layer.
We term the task-specific gating network with a short name ``{{task gating}}'', and denote it as $f_g^{l,t}(\cdot): \mathbb{R}^{N\times C \rightarrow N\times K}$, where $t\in [1,T]$ ($T$ is the total number of tasks) is a task index. The task-gating scores $\mG^{l,t} \in \mathbb{R}^{N\times K}$
is computed as $\mG^{l,t} = f_g^{l,t}(\mX^l)$.

\par For the design of the gating networks, previous Mixture-of-Experts methods~\cite{fedus2021switch,lepikhin2020gshard,shazeer2017outrageously} use a simple linear layer followed by a Softmax function, which does not take into account the context of each image patch token, while using only the corresponding patch token when computing the gating score.
As various challenging computer vision tasks~(\eg semantic segmentation and monocular depth estimation) require modeling and understanding each image patch with context information, we thus design a context-aware gating strategy.
Specifically, context-aware gating adopts convolution kernels to extend the receptive field of the gating networks, and thus it can involve a local context of tokens in the computation of gating scores. We first reshape the input token sequence $\mX$~(in $\mathbb{R}^{N\times C}$) as spatial maps (in  $\mathbb{R}^{H\times W \times C}$), and then compute the gating scores of $K$ experts with three layers, each having a 3$\times$3 convolution with ReLU as intermediate activation function. We denote this computation procedure as ``$\mathrm{CONV}$'', which will also be used later in Equation~\ref{eq:gating}.

\subsubsection{Multi-Task Feature Memory}
\label{sec:task_memory}
To conduct long-range modeling of the multi-task representations decoded at different network layers that contain different levels of visual information, we design a multi-task feature memory to interact with the MMoE modules through read and write operations. The read operation uses the task feature memory as an additional feature expert for task-specific feature decoding in MMoE. The write operation updates the multi-task feature memory by using the decoded task-specific features from MMoE. In this way, the multi-task feature memory can aggregate the decoded task-specific features crossing different network layers.
Specifically, the read operation is an identity mapping. The write operation can be performed as follows: given a decoded task-specific feature of the $t$-th task at the first layer, \ie~$\mF^{1,t}$, the initialization of the multi-task feature memory $\mM^{1,t}$ is obtained by $\mM^{1,t} \leftarrow \mF^{1,t}$.
In the subsequent layers,
suppose we decode task-specific features at every network layer, at the $l$-th layer ($l>1$), we write the multi-task feature memory~$\{\mM^{l,t}, t\in [1,T] \}$ with the decoded task-specific feature $\mF^{l,t}$ weighted by a task-specific learnable momentum value $\alpha^{l,t}$ as shown in Figure~\ref{fig:mmoe}:
\begin{equation}
    \mM^{l,t} \leftarrow \mF^{l,t} + \alpha^{l,t} \mM^{l-1,t}, \ \ l > 1.
\end{equation}

\subsubsection{Dynamically Assembling Task-Specific Features}
With the obtained representative task-generic features and task-specific gating scores as above-described, we can perform dynamic decoding of task-specific features, as the gating parameters are predicted in a sample-dependent manner.~This is significantly different from existing decoder-focused multi-task structures, which share the decoding parameters for all the different input samples.
Specifically, for the $t$-th task, we multiply every token of the representative task-generic features~$\{\mR_i^l,i\in[1,K]\}$ with their corresponding task-gating score~$\{\mG^{l,t}_i,i\in[1,K]\}$ and sum the tokens in the same spatial position to decode task-specific feature.
Furthermore, to learn more discriminative features for each task, we utilize the corresponding task-specific feature in multi-task feature memory as an additional feature expert in MMoE, which contains long-range task-specific information learned throughout the entire network. It helps decode more effective and long-term task-specific features.
As shown in Figure~\ref{fig:mmoe}, for the MMoE at the $l$-th transformer layer ($l>1$), we compute an additional gating score map $\mG^{l,t}_{K+1}$ for the task-specific feature from the multi-task feature memory $\mM^{l-1,t}$, which is performed by {adding one more output channel into the task gating network} $f_g^{l,t}: \mathbb{R}^{N\times C}  \rightarrow \mathbb{R}^{N\times {(K+1)}}$.
In summary, the task gating scores of the $t$-th task and the $l$-th layer, \ie,
~$\mG^{l,t} = \{ \mG^{l,t}_1, \mG^{l,t}_2,..., \mG^{l,t}_{K+1} \}$,
is computed as:
\begin{equation}
\begin{aligned}
\label{eq:gating}
\end{aligned}
\mG^{l,t}=\left\{
\begin{array}{ll}
\textrm{Softmax} (\mathrm{CONV}^{K+1}  (\mX^l))  & l>1,\\
\textrm{Softmax} (\mathrm{CONV}^{K}  (\mX^l))  &l=1,\\
\end{array} \right.
\end{equation}
where the superscript $K$ of $\mathrm{CONV}$ denotes the number of output convolution kernels.
And accordingly, the decoded task-specific features at the $l$-th MMoE module is:
\begin{equation}
\mF^{l,t}=\left\{
\begin{array}{ll}
\sum_{i=1}^{K} \mG^{l,t}_i \times \mR_i^l + \mG^{l,t}_{K+1}\times \mM^{l-1,t}  & l>1,\\
\sum_{i=1}^{K} \mG^{l,t}_i \times \mR_i^l  &l=1.\\
\end{array} \right.
\label{eq:task_fea}
\end{equation}

\par\noindent \textbf{Alternative sparse activation of experts}
As an alternative, we can also use ``sparse connections'' in MMoE, which means that we only activate $K_{sel}$ selected experts to compute the task-specific features ($K_{sel}<K$) for each token~\cite{shazeer2017outrageously}. In this case, for each token, we find the top $K_{sel}$ experts to be activated based on the task-gating scores, and sum their outputs after being weighted by the task-gating scores.
A comparison between the dense connection and sparse connection in MMoE is discussed in the experiments.

\subsection{Generation of Multi-Task Predictions}
At the final network layer~$L$, we use task-specific prediction heads to directly generate prediction outputs for each task from the multi-task feature memory~~$\{\mM^{L, t}, t\in [1, T]\}$. The prediction heads for different tasks have a similar structure, consisting of a 3$\times$3 convolution layer with batch normalization and ReLU activation, followed by a linear layer for the final prediction.

\section{Experiment}
\label{sec:exps}
In this section, we validate the effectiveness of our proposal through several aspects: (i) ablation study, (ii) comparison with SOTA methods, and (iii) qualitative analysis.

\begin{figure}[!t]
    \centering
     \includegraphics[width=1.\linewidth]{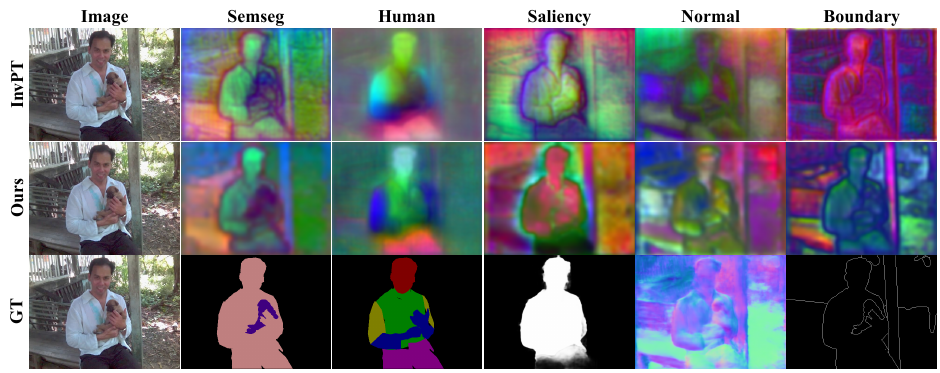}
     \vspace{-20pt}
     \caption{{PCA visualization of the decoded final task features generated by InvPT~\cite{invpt2022} and our TaskExpert. Our model generates more discriminative features for all tasks.}}
     \label{fig:vis_task_feas_pca}
\vspace{-10pt}
\end{figure}

\subsection{Experimental Setup}
\label{sec:setup}
\par\noindent\textbf{Tasks and Benchmarks} For evaluation, we adopt two popular benchmarks in multi-task learning as previous works~\cite{invpt2022,mti}, \ie~PASCAL-Context~\cite{chen2014detect,everingham2010pascal} and NYUD-v2~\cite{silberman2012indoor} datasets. PASCAL-Context contains 4,998 samples in train split and 5,105 samples in test split. It provides labels for multiple important visual understanding tasks, including semantic segmentation, human parsing, and object boundary detection. Previous work on multi-task learning generates pseudo labels for another two tasks~\cite{astmt}: surface normals estimation and saliency detection.
Meanwhile, NYUD-v2 contains 1,449 samples in train split and 795 samples in test split.  It collects labels for  semantic segmentation and monocular depth estimation, from which the labels of surface normal estimation and object boundary detection can be generated.
We utilize all the tasks provided in these datasets for evaluation.

\par\noindent\textbf{Metrics}
We use the same evaluation metrics as previous works~\cite{invpt2022}.
Specifically, semantic segmentation (Semseg) and human parsing (Parsing) tasks are evaluated by the mean Intersection over Union (mIoU). Monocular depth estimation task (Depth)  is evaluated by Root Mean Square Error (RMSE).
Surface normal estimation task (Normal) is evaluated by the mean error (mErr) of angles. Saliency detection task (Saliency) is evaluated by maximal F-measure~(maxF). Object boundary detection task~(Boundary) is evaluated by the optimal-dataset-scale F-measure (odsF). We calculate the mean relative difference across all tasks compared to the single-task baseline, which is noted as Multi-task Performance~(MTL Perf $\Delta_m$)~\cite{astmt}.

\par\noindent\textbf{Implementation Details}
We use 5 expert networks in MMoE for PASCAL-Context and 4 expert networks for NYUD-v2, which are the same as the corresponding number of tasks. We add MMoE at the end of each transformer layer if not otherwise stated.
For model training, we follow the setting of InvPT~\cite{invpt2022} and use Adam optimizer to train the models for 40,000 iterations with a batch size of 4. Polynomial learning rate scheduler is used with a learning rate of $2\times 10^{-5}$ and a weight decay rate of $1\times 10^{-6}$.
We adopt the loss functions and weights for different tasks in \cite{invpt2022}.
The ViT backbones are pre-trained on ImageNet-22K~\cite{deng2009imagenet}.

\begin{table}[t]
\centering
\caption{Ablation study on PASCAL-Context with a ViT-Base backbone. ``MoE'' denotes using Mixture-of-Experts module; ``CG'' denotes using context-aware gating; ``Mem'' denotes using our multi-task feature memory. $\mathbf{\downarrow}$' indicates lower better and `$\mathbf{\uparrow}$' means higher better.}
\label{tab:modules}
\vspace{-10pt}
\resizebox{1\linewidth}{!}{
    \begin{tabular}{l|cccccc}
    \toprule
        \multicolumn{1}{c|}{ \multirow{2}*{ \textbf{Model}} } & \textbf{Semseg}  & \textbf{Parsing}  & \textbf{Saliency} & \textbf{Normal} & \textbf{Boundary}&  \textbf{MTL Perf}   \\
        & mIoU $\mathbf{\uparrow}$  & mIoU $\mathbf{\uparrow}$
      & maxF $\mathbf{\uparrow}$ & mErr $\mathbf{\downarrow}$ & odsF $\mathbf{\uparrow}$ & $\Delta_m$ $\mathbf{\uparrow}$ \\
    \midrule
    Baseline & 76.40 & 65.70 & 84.63 & 13.83 & 69.50 & -3.70 \\
    w/ MoE & 77.62 & 66.37 & 84.65 & 13.77 & 70.40 &-2.88 \\
     w/ MoE+CG & 77.99 & 66.82 & 84.76 & 13.64 & 71.70& -2.07\\
     w/ MoE+CG+Mem & \textbf{78.45} & \textbf{67.38} & \textbf{84.96} & \textbf{13.55} & \textbf{72.30} & \textbf{-1.46} \\
    \bottomrule
    \end{tabular}}
     \vspace{-20pt}
\end{table}

\subsection{Ablation Study}
\label{sec:ablation}

\par\noindent \textbf{Model Variants Definition}
To examine the effectiveness of TaskExpert under a fair setting, we compare it to the multi-task baseline model and its variants with the same backbone.
(1) The baseline model (``Baseline'') uses ViT-Base as backbone, and designs task-specific decoders in each transformer layer to generate layer-wise task-specific features. The task-specific features from all layers are added together as multi-scale input of the corresponding prediction head for each task. These layer-wise decoders share the same structure as the expert networks used in MMoE.
The prediction heads for all the tasks have a similar structure:  Conv(3$\times$3)-BN-ReLU-Linear, which are the same as those used in TaskExpert.
(2) ``TaskExpert w/ MoE'' adds the expert and gating networks of MMoE without context-aware gating to Baseline. It uses gating networks to control the contribution of each expert to each task in a sample-dependent manner without context-aware gating or multi-task feature memory. The number of experts is the same as the number of tasks.
(3) ``TaskExpert w/ MoE+CG'' further uses our context-aware gating strategy, which improves the gating networks by introducing the contextual information of each token.
(4) ``TaskExpert w/ MoE+CG+Mem'' is the full version of our TaskExpert using multi-task feature memory in the proposed MMoE module to establish modeling of long-term task-specific representation.

\begin{figure}[!t]
    \centering
     \includegraphics[width=1.\linewidth]{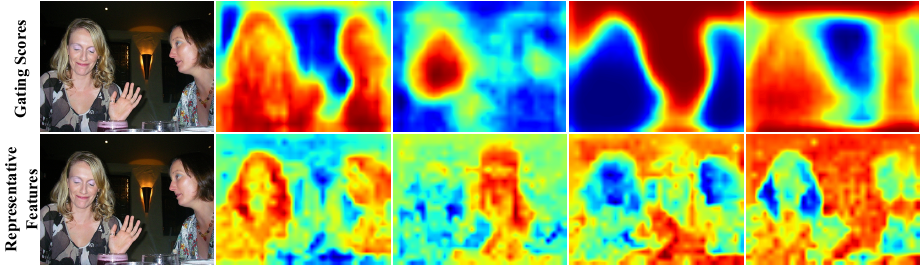}
     \vspace{-20pt}
     \caption{Visualization of the task-gating scores by task-specific gating networks and the decomposed representative features from expert networks in MMoE modules. The gating scores for different experts show distinct distributions, while the representative features are discriminative to different visual elements. These results suggest different experts are able to decompose diverse visual information from the backbone feature.}
     \label{fig:vis_path_rep}
\vspace{-10pt}
\end{figure}

\par\noindent \textbf{Key Results}
We analyze the performances of different model variants introduced above in  Table~\ref{tab:modules}. We find that using MoE brings a clear performance gain compared to the baseline model, which demonstrates the effectiveness of designing task-specific gating networks to aggregate the representative task-generic features for dynamically decoding task-specific features. For example, mIoU of semantic segmentation is improved from 76.40 to 77.62~(+1.22), and human parsing is improved from 65.70 to 66.37~(+0.67).
Furthermore, the usage of context-aware gating strategy helps MoE achieves better performance by enabling the learning of contextual information for each token when computing the gating scores. For example, odsF of boundary detection is improved from 70.40 to 71.70~(+1.30) compared to only using MoE.
Finally, with long-term modeling of task-specific representations via the proposed multi-task feature memory of Memorial Mixture-of-Experts, TaskExpert achieves significant performance improvement compared to the baseline model on different tasks, and the overall multi-task performance $\Delta_m$ is improved by \textbf{+2.24}.

\begin{figure}[!t]
    \centering
     \includegraphics[width=1\linewidth]{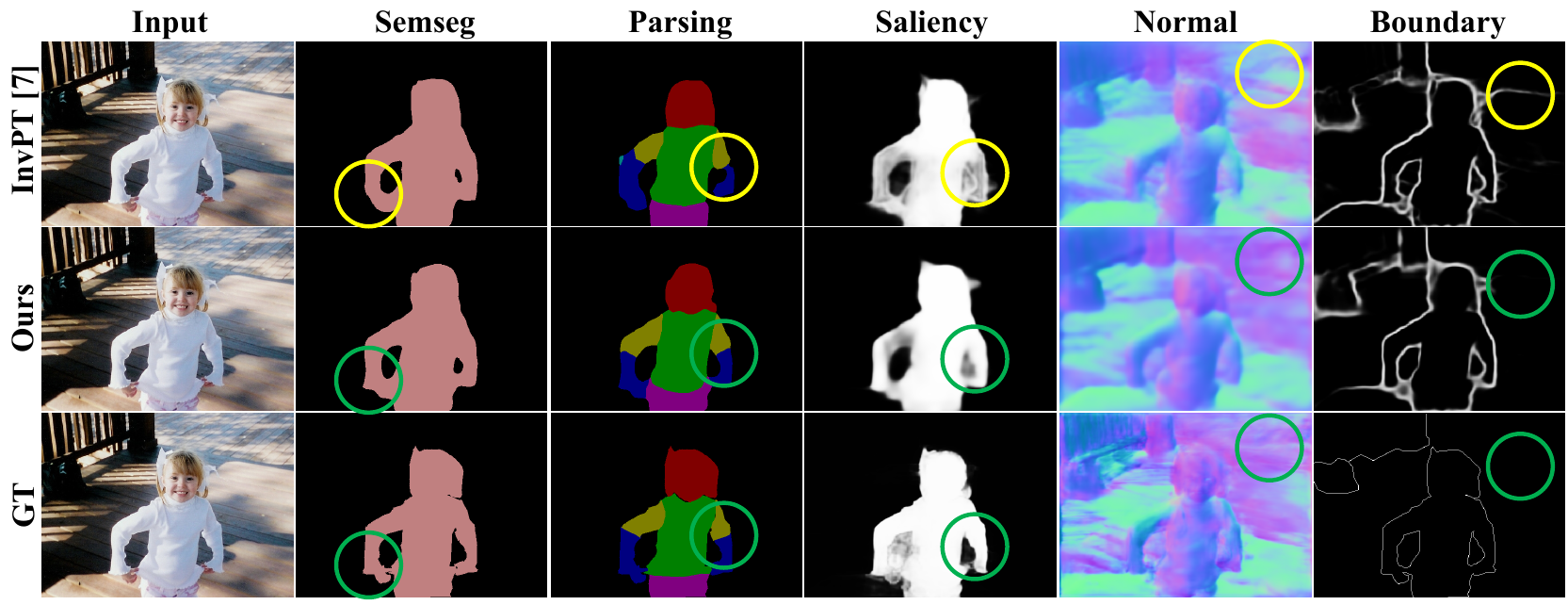}
     \vspace{-18pt}
     \caption{Qualitative comparison with SOTA method (\ie InvPT~\cite{invpt2022}) on PASCAL-Context. Our TaskExpert generates better predictions with more accurate details as highlighted. }
     \label{fig:qualitative_pascal}
\vspace{-10pt}
\end{figure}

\par
\begin{figure*}[!t]
    \centering
    \vspace{-20pt}
\includegraphics[width=1\linewidth]{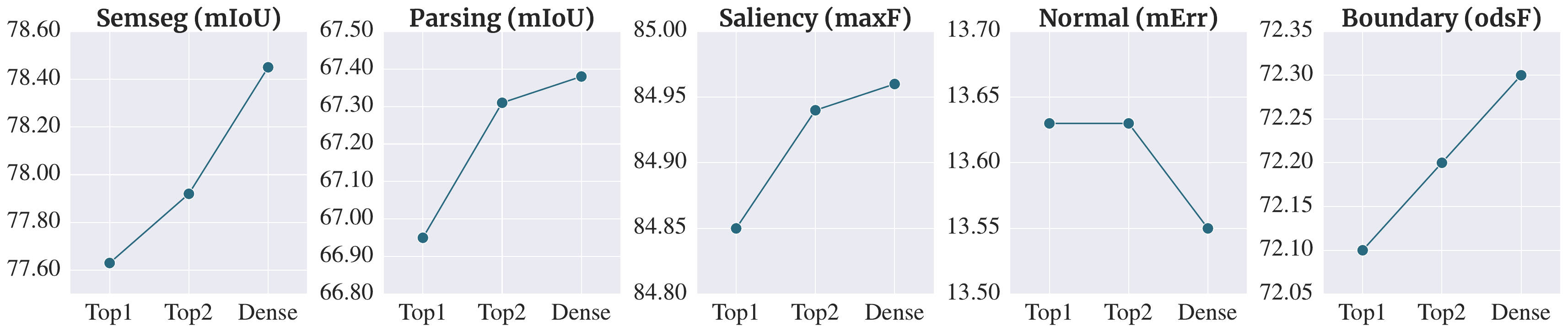}
\vspace{-20pt}
\caption{Comparison of TaskExpert with dense and sparse connection in MMoE. ``Top 1'' denotes only activating the expert with the highest gating score, ``Top 2'' means activating the experts with top 2 gating scores, and ``Dense'' activates all experts.}
     \label{fig:activate_expert_no}
\vspace{-10pt}
\end{figure*}

\noindent\textbf{Study of Spare and Dense Connections in MoE}
The original MoE models use the dense connection as the gating mechanism: all experts are activated and contribute to the output~\cite{jacobs1993learning,jacobs1991adaptive}. Recent works propose to utilize sparse connection as the gating mechanism: for each token, only a subset of experts contribute to the output~\cite{shazeer2017outrageously}, as explained in Section~\ref{sec:mmoe}.
In this experiment, we study the performances of using sparse connection (top 1 and top 2), and compare them with the default dense connection manner on PASCAL-Context. The results are demonstrated in Figure~\ref{fig:activate_expert_no}.
It can be observed that activating more experts leads to better performance, and the dense connection achieves the best results. The reason is that it is able to freely utilize the full representative ability of all expert networks.

\par\noindent\textbf{Light-Weight Versions of TaskExpert}
To study the model performance and efficiency with a smaller parameter and memory budget,  we conduct 2 sets of experiments in Table~\ref{tab:fewer_moem}:
\textbf{(i)} Using different numbers of MMoE modules on ViT-B backbone for PASCAL-Context. The MMoE modules are embedded evenly along the backbone's depth.
\textbf{(ii)} Implementing TaskExpert on ViT-S, which is a smaller backbone.
These smaller variants achieve reasonable performance with less parameter and  memory consumption.

\begin{table}[!t]
\huge
\centering
\vspace{-5pt}
\caption{Variants of TaskExpert on PASCAL-Context. The inference GPU memory is measured with batch size 1.}
\label{tab:moe}
\vspace{-10pt}
\resizebox{0.99\linewidth}{!}{
    \begin{tabular}{c|cccccccc}
    \toprule
        \multicolumn{1}{c|}{ \multirow{2}*{ \textbf{Variant}} } & \textbf{Semseg}  & \textbf{Parsing}  & \textbf{Saliency} & \textbf{Normal} & \textbf{Boundary} & \multirow{2}*{ \textbf{\#Param}} & \multirow{2}*{ \textbf{FLOPS}} & \multirow{1}*{ \textbf{Inference}}  \\
        & mIoU $\mathbf{\uparrow}$  & mIoU $\mathbf{\uparrow}$
       & maxF $\mathbf{\uparrow}$ & mErr $\mathbf{\downarrow}$ & odsF $\mathbf{\uparrow}$ & & &\textbf{GPU Mem} \\
    \midrule
   3 MMoE & 78.04 & 66.54 & 84.85 & 13.62 & 71.20 & 174M & 597G & 8367M \\
   6 MMoE & 78.18 & 66.79 & 84.93 & 13.59 & 71.70 & 234M & 659G & 8851M \\
    12 MMoE & \textbf{78.45} & \textbf{67.38} & \textbf{84.96} & \textbf{13.55} & \textbf{72.30} & 347M & 782G & 12503M \\
    \midrule
    ViT-S & 75.04 & 62.68 & 84.68 & 14.22 & 68.80 & \textbf{55M} & \textbf{204G} & \textbf{4187M} \\
   ViT-B & \textbf{78.45} & \textbf{67.38} & \textbf{84.96} & \textbf{13.55} & \textbf{72.30} & 347M & 782G  & 12503M\\
    \bottomrule
    \end{tabular}}
     \vspace{-10pt}
\end{table}

\begin{table}[!t]
\huge
\centering
\caption{Comparison with multi-task MoE methods with ViT-S backbone on PASCAL-Context. The training GPU time is shown in the last column.}
\label{tab:fewer_moem}
\vspace{-10pt}
\resizebox{0.99\linewidth}{!}{
    \begin{tabular}{c|cccccccc}
    \toprule
        \multicolumn{1}{c|}{ \multirow{2}*{ \textbf{Variant}} } & \textbf{Semseg}  & \textbf{Parsing}  & \textbf{Saliency} & \textbf{Normal} & \textbf{Boundary} & \multirow{2}*{ \textbf{\#Param}} &  \multirow{2}*{ \textbf{FLOPS}} & \multirow{1}*{\textbf{GPU}}  \\
        & mIoU $\mathbf{\uparrow}$  & mIoU $\mathbf{\uparrow}$
       & maxF $\mathbf{\uparrow}$ & mErr $\mathbf{\downarrow}$ & odsF $\mathbf{\uparrow}$ && & \multirow{1}*{\textbf{Hours}} \\
    \midrule
    M$^3$ViT~\cite{m3vit} & 72.80 & 62.10 & 66.30 & 14.50 & 71.70 & \textbf{42M} & {420G} & $\sim$ 72 \\
    Mod-Squad~\cite{chen2022mod} & 74.10 & \textbf{62.70} & 66.90 & \textbf{13.70} & \textbf{72.00} & 50M & {420G} & $\sim$ 72\\
     Ours & \textbf{75.04} & 62.68 & \textbf{84.68} & 14.22 & 68.80 & 55M & \textbf{204G} & \textbf{$\sim$ 5}\\
    \bottomrule
    \end{tabular}}
     \vspace{-10pt}
\end{table}

\begin{table}[t]
\centering\caption{Comparison with the state-of-the-art methods with the same backbone on PASCAL-Context. Our TaskExpert clearly achieves superior performances on \textit{all} tasks. }
\vspace{-10pt}
\centering
\resizebox{1.0\linewidth}{!}{
\setlength{\tabcolsep}{1mm}{
    \begin{tabular}{l|ccccccc}
    \toprule
   \multicolumn{1}{c|}{\multirow{2}*{ \textbf{Model} }} & \multirow{2}*{\textbf{FLOPs}} & \multirow{2}*{\textbf{\#Param}}  & \textbf{Semseg}  & \textbf{Parsing}
       & \textbf{Saliency} & \textbf{Normal} & \textbf{Boundary}
     \\
        &&& mIoU $\mathbf{\uparrow}$  & mIoU $\mathbf{\uparrow}$
       & maxF $\mathbf{\uparrow}$ & mErr $\mathbf{\downarrow}$ & odsF $\mathbf{\uparrow}$
     \\
    \midrule
    PAD-Net~\cite{padnet} & 124G & 81M  & 53.60 & 59.60 & 65.80 & 15.30 & 72.50\\
    MTI-Net~\cite{mti} & 161G & 128M &  61.70 & 60.18 & 84.78 & 14.23 & 70.80\\
    ATRC~\cite{atrc} & 216G & 96M& 67.67 & 62.93 & 82.29 & 14.24 & 72.42\\
    \midrule
    PAD-Net w/ ViT-L~\cite{padnet} &  773G   &  330M  &  78.01  &  67.12  &  79.21  &  14.37  &  72.60 \\
    MTI-Net w/ ViT-L~\cite{mti} & 774G  &   851M   &  78.31   &  67.40   &  84.75   & 14.67   & 73.00  \\
    ATRC w/ ViT-L~\cite{atrc} &871G  &  340M   &  77.11   &  66.84   &  81.20   & 14.23  &  72.10  \\
    InvPT~\cite{invpt2022}& 669G& 423M & 79.03 & 67.61 & 84.81 & 14.15 & 73.00\\
    TaskExpert (\textbf{ours})&622G& 420M& \textbf{80.64} &\textbf{69.42} & \textbf{84.87} & \textbf{13.56} & \textbf{73.30}\\
    \bottomrule
    \end{tabular}}}
    \label{tab:sota_pascal}
    \vspace{-8pt}
\end{table}

\subsection{Comparison with SOTA Methods}
\label{sec:sota}

We implement TaskExpert based on ViT-Large backbone with MMoE modules embedded at every two transformer layers, then we compare TaskExpert to recent SOTA methods in literature on both PASCAL-Context and NYUD-v2 in Table~\ref{tab:sota_pascal} and Table~\ref{tab:sota_nyud}.
We re-implement several CNN-based SOTA multi-task learning methods on ViT-L backbone for a more comprehensive comparison on PASCAL-Context.
Compared with the previous SOTA model (\ie InvPT~\cite{invpt2022}), TaskExpert achieves clearly better performances on \textbf{\textit{all 9 metrics}} of these two challenging benchmarks with the same backbone.
Particularly, on PASCAL-Context, we improve semantic segmentation by \textbf{+1.61} and human parsing by \textbf{+1.81}. On NYUD-v2, semantic segmentation is improved by \textbf{+1.79}.~It should be noted that, our gains are measured upon the very strong performance foundation established by InvPT (\emph{e.g.}~Semseg~79.03, Saliency~84.81 on PASCAL-Context).
This superior performance clearly shows the effectiveness of the proposed TaskExpert.
TaskExpert also maintains a \textit{\textbf{higher efficiency}} with a reduction of 3 million in the number of parameters and 47 billion FLOPs compared to InvPT.

As underscored in the Related Work section, our TaskExpert is designed for a different setting from  M$^3$ViT~\cite{m3vit} and Mod-Squad~\cite{chen2022mod}. Their models are constrained to generating a single task prediction per forward pass, while our model capably generates predictions for all tasks within a single forward pass.
{Therefore, the multi-task learning efficiency of our model is significantly higher than their methods.
When utilizing the same backbone, it requires \textit{only half the FLOPS}, as demonstrated in Table~\ref{tab:fewer_moem}.
Moreover, our model's training duration is markedly shorter since it learns all tasks concurrently in each iteration.
While they necessitate around 18 hours of training with 4 A6000 GPUs as confirmed by their authors, our model completes training in under 5 hours using a single NVIDIA 3090 GPU.
}

\begin{table}[!t]
\centering\caption{Comparison with the state-of-the-art methods on NYUD-v2 dataset. Our TaskExpert achieves the best performances on \textit{all} tasks. }
\vspace{-10pt}
\centering
\resizebox{1.\linewidth}{!}{
\setlength{\tabcolsep}{1.5mm}{
    \begin{tabular}{l|cccccc}
    \toprule
         \multicolumn{1}{c|}{\multirow{2}*{ \textbf{Model} }} & \multirow{2}*{\textbf{Backbone}} & \textbf{Semseg}  & \textbf{Depth}  & \textbf{Normal} & \textbf{Boundary} \\
        & &mIoU $\mathbf{\uparrow}$ & RMSE $\mathbf{\downarrow}$ & mErr $\mathbf{\downarrow}$ & odsF $\mathbf{\uparrow}$ \\
    \midrule
    Cross-Stitch~\cite{crossstitch} & HRNet18 & 36.34 & 0.6290 & 20.88 &76.38    \\
    PAP~\cite{papnet} & HRNet18 & 36.72 & 0.6178 &20.82 & 76.42 \\
    PSD~\cite{psd}& HRNet18 & 36.69 & 0.6246 & 20.87 & 76.42 \\
    PAD-Net~\cite{padnet}& HRNet18 & 36.61 & 0.6270 & 20.85 & 76.38\\
    MTI-Net~\cite{mti}& HRNet48 & 45.97  & 0.5365 & 20.27 & 77.86 \\
    ATRC~\cite{atrc}& HRNet48  & 46.33  & 0.5363 & 20.18 & 77.94 \\
    InvPT~\cite{invpt2022}& ViT-L  &53.56 & 0.5183 & 19.04 & 78.10\\
    TaskExpert (\textbf{ours})& ViT-L& \textbf{55.35} & \textbf{0.5157} & \textbf{18.54} & \textbf{78.40}  \\
    \bottomrule
    \end{tabular}}}
    \label{tab:sota_nyud}
    \vspace{-16pt}
\end{table}

\subsection{Qualitative study}
\label{sec:qualitative}

\par\noindent\textbf{Qualitative Study of MMoE}
To understand better the learned gating scores and representative task-generic features by the gating networks and expert networks in our MMoE modules, we present qualitative examples in Figure~\ref{fig:vis_path_rep}.
As can be observed, TaskExpert effectively learns representative task-generic features with distinct activation distributions, focusing on different aspects of the image.
Accordingly, the gating scores exhibit different response intensities to different areas for selectively assembling the experts' features for decoding task-specific features.

\par\noindent\textbf{PCA Visualization of Representative Features of Task-Experts at Different Layers} {
We conduct principal component analysis~(PCA) on the representative features at layers 1, 6, and 12. As shown in Fig.~\ref{fig:vis_rep_features_layers}, we visualize the first three components of the feature PCA as a color map. It reveals that the task experts at deeper layers capture more high-level semantic information of the input. On the other hand, in the shallower layers, the task-feature experts focus more on low-level visual patterns (\eg, edges).
}

\par\noindent\textbf{Qualitative Comparison of Decoded Task Features} {
In Fig.~\ref{fig:vis_task_feas_pca}, we also conduct PCA on the final task features generated by InvPT~\cite{invpt2022} and our TaskExpert, and draw the first three components. The features for different tasks from TaskExpert show clear task-specific characteristics and are much more discriminative than those of InvPT.
}

\par\noindent\textbf{Qualitative Comparison of Prediction Results}
To intuitively examine the prediction quality of TaskExpert compared to the previous best-performing method~\cite{invpt2022}, we visualize the predictions of the two models on the test split of PASCAL-Context in Figure~\ref{fig:qualitative_pascal}.
Our TaskExpert generates visually better multi-task predictions.

\section{Conclusion}
In this paper, we present TaskExpert, a novel multi-task mixture-of-expert model establishing new SOTA performances on two challenging benchmarks.
TaskExpert can effectively perform dynamic decoding of task-specific features, from a set of representative task-generic features learned by expert networks.
TaskExpert further models the long-term task-specific representations at different layers via a designed multi-task feature memory.
Extensive experiments clearly prove the effectiveness and efficiency of our TaskExpert model, and we hope that our TaskExpert can also be beneficial for the future study of dynamic representation learning problems.

\clearpage

\section{Additional Implementation Details}
This section provides additional implementation details regarding the data processing and the model optimization for our proposed model.

\par\noindent\textbf{Data Processing}
We follow the data processing pipeline of InvPT~\cite{invpt2022} for a fair comparison. Specifically, on NYUD-v2 dataset, we crop the input image to the size of $448\times 576$ randomly. On PASCAL-Context dataset, we pad the image to a size of $512\times 512$. We also use the same data augmentation techniques, including random scaling, color jittering, cropping, and horizontal flipping.

\par\noindent\textbf{Model Optimization}
We totally investigate 6 different tasks on the challenging NYUD-v2~\cite{silberman2012indoor} and PASCAL-Context~\cite{chen2014detect} benchmarks, including semantic segmentation~(Semseg), monocular depth estimation~(Depth), surface normal estimation~(Normal), human parsing~(Parsing), saliency detection~(Saliency), and object boundary detection~(Boundary).
For the continuous regression tasks, such as Depth and Normal, we employ the $\mathcal{L}1$ Loss. For the discrete classification tasks, including Semseg, Parsing, Saliency, and Boundary, we utilize the cross-entropy loss. The overall loss is a weighted sum of the task losses from all the tasks, using a loss weight dictionary utilized by~\cite{invpt2022,mti}.
More specifically, the loss weight is 1 for Semseg, 2 for Parsing, 5 for Saliency, 50 for Boundary, 1 for Depth, and 10 for Normal.

\begin{table}[h]
\centering
\caption{Quantitative comparison of using different sizes of convolution kernels in the gating networks. $3\times3$ is large enough to perceive the context of each token for gating networks. It clearly outperforms the variant with $1\times 1$ kernel on all the tasks, and achieves the best performance on most of the tasks compared to all the variants.}
\vspace{-10pt}
\label{tab:kernel_size}
\resizebox{1\linewidth}{!}{
    \begin{tabular}{c|cccccc}
    \toprule
        \multicolumn{1}{c|}{ \multirow{2}*{ \textbf{Kernel Size}} } & \textbf{Semseg}  & \textbf{Parsing}  & \textbf{Saliency} & \textbf{Normal} & \textbf{Boundary}   \\
        & mIoU $\mathbf{\uparrow}$  & mIoU $\mathbf{\uparrow}$
      & maxF $\mathbf{\uparrow}$ & mErr $\mathbf{\downarrow}$ & odsF $\mathbf{\uparrow}$ \\
    \midrule
    1$\times$1 & 77.73 & 66.43 & 84.67 & 13.78 & 71.00\\
    3$\times$3 & \textbf{78.45} & \textbf{67.38} & 84.96 & 13.55 & \textbf{72.30} \\
    5$\times$5 & 78.40 & 66.59 & \textbf{85.22} & \textbf{13.48} & 72.10\\
    \bottomrule
    \end{tabular}}
     \vspace{-10pt}
\end{table}

\section{More Study of Gating Networks}
\subsection{Ablation Study of Kernel Sizes in Context-Aware Gating Networks}
Context-Aware Gating is a critical module in MMoE that incorporates contextual information into the computation of the gating score for each token.
In this experimental study, we aim to study the influence of kernel sizes used in convolutions of the gating networks in Table~\ref{tab:kernel_size}. Compared with the variant using $1\times 1$ convolution kernel size~which resembles an MLP, {using $3\times 3$ clearly brings improvement on the performances of all the tasks.} For instance, the mIoU of Semseg and Parsing is improved from 77.73 and 66.43 to 78.45~(+0.72) and 67.38~(+0.95), respectively.
However, further increasing the kernel size does not lead to further improvement on all the tasks. This is because the most important information for computing the gating score of a given token typically comes from spatially adjacent tokens. For this reason, we have chosen to use a default convolution kernel size of $3\times 3$ in our context-aware gating module.

\subsection{Comparison with MLP Gating Network}
A standard Mixture-of-Experts (MoE) gating network typically incorporates a simplistic linear layer. To establish a balanced comparison, ensuring a comparable parameter size, we implement a conventional Multi-Layer Perceptron (MLP) consisting of two linear layers as the gating network, without incorporating the multi-task feature memory mechanism.
The corresponding results, presented in Table~\ref{tab:mlpgate}, highlight that our context-aware gating (CG) strategy achieves superior performance, while simultaneously consuming fewer parameters.

\begin{table}[t]
\centering
\caption{Ablation study of gating networks on PASCAL-Context. ``CG'' denotes the proposed context-aware gating.}
\label{tab:mlpgate}
\vspace{-10pt}
\resizebox{1\linewidth}{!}{
    \begin{tabular}{c|ccccccc}
    \toprule
        \multicolumn{1}{c|}{ \multirow{2}*{ \textbf{Model}} } & \textbf{Semseg}  & \textbf{Parsing}  & \textbf{Saliency} & \textbf{Normal} & \textbf{Boundary}&  \textbf{MTL Perf}  & \multirow{2}*{ \textbf{\#Param}}   \\
        & mIoU $\mathbf{\uparrow}$  & mIoU $\mathbf{\uparrow}$
      & maxF $\mathbf{\uparrow}$ & mErr $\mathbf{\downarrow}$ & odsF $\mathbf{\uparrow}$ & $\Delta_m$ $\mathbf{\uparrow}$  \\
    \midrule
    MLP-Gating & 77.13 & 66.25 & 84.64 & 13.80 & 70.80 & -2.96 & 256M  \\
    CG & 77.99 & 66.82 & 84.76 & 13.64 & 71.70& -2.07 & 227M \\
    \bottomrule
    \end{tabular}}
\end{table}

\section{More Comparison with Previous SOTA}
\begin{figure*}[!t]
    \centering
     \includegraphics[width=.95\linewidth]{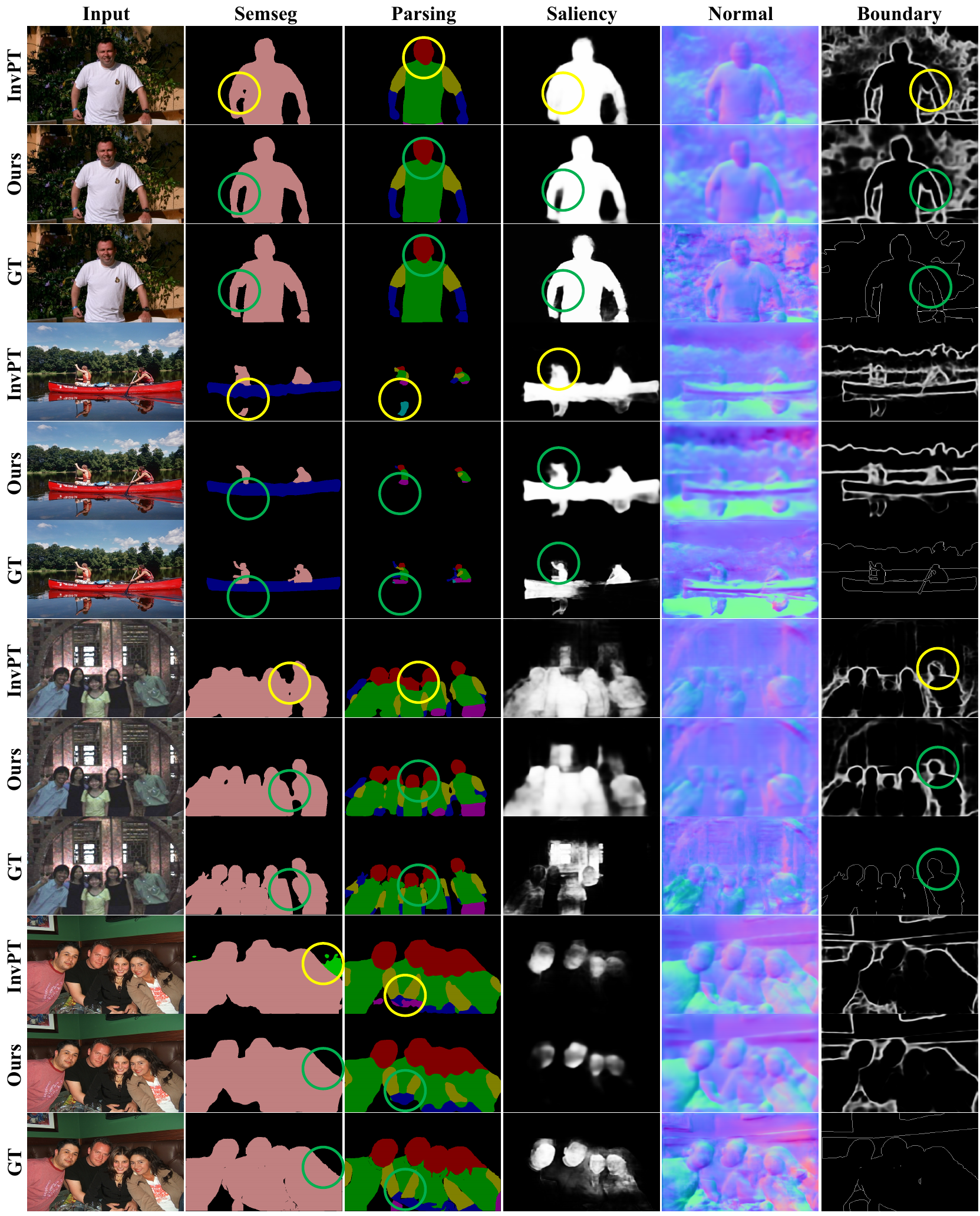}
     \vspace{-10pt}
     \caption{Qualitative comparison with previous best-performing method InvPT~\cite{invpt2022} on PASCAL-Context. Our method generates significantly better results, especially on semantic segmentation and human parsing, as highlighted in circles.}
     \label{fig:qualitative_pascal2}
\end{figure*}

\begin{figure*}[!t]
    \centering
    \includegraphics[width=.7\linewidth]{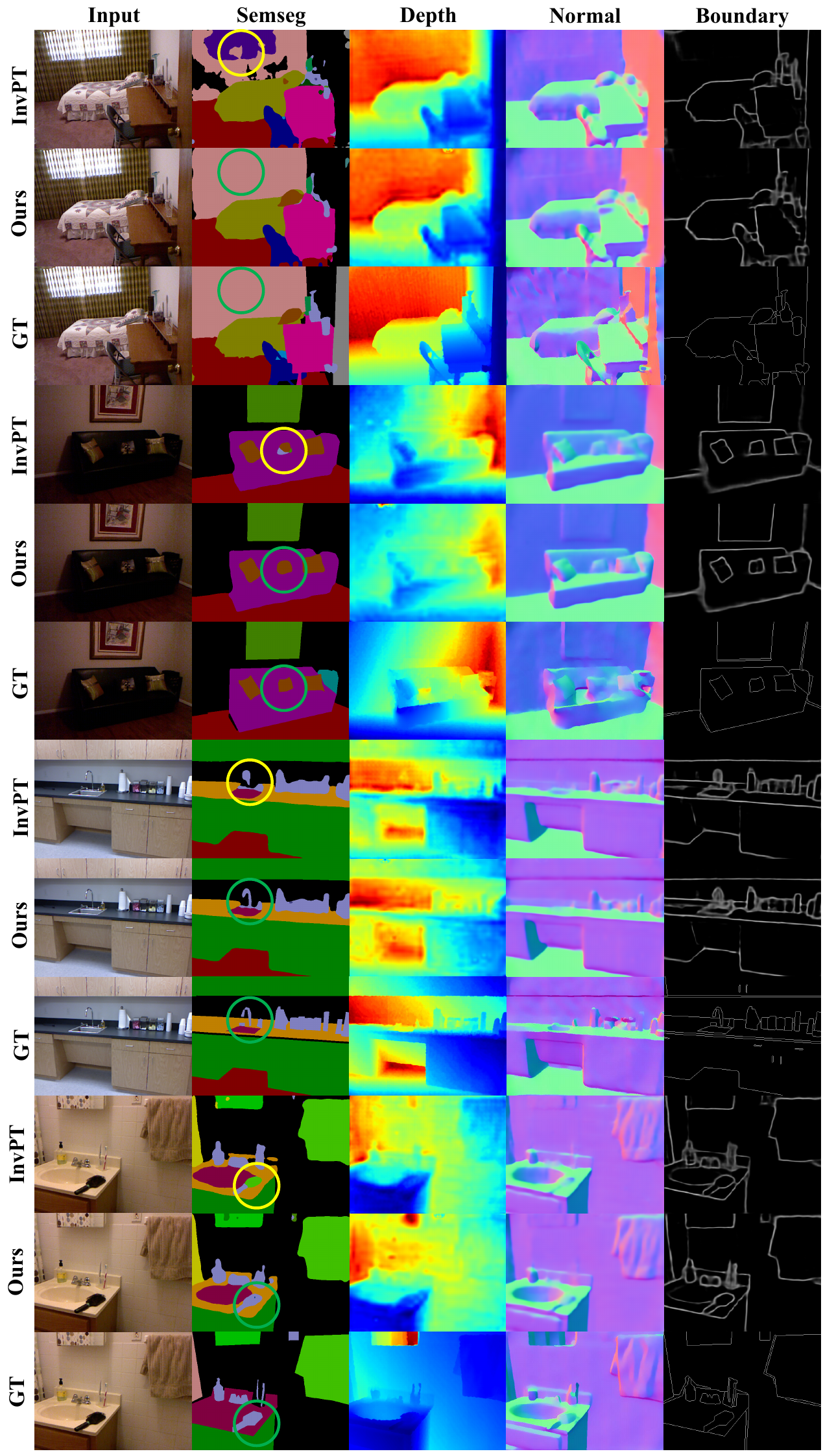}
     \vspace{-10pt}
     \caption{Qualitative comparison with the previous best-performing method InvPT~\cite{invpt2022} on NYUD-v2. Our method generates significantly better results than previous best-performing InvPT, as highlighted in circles.}
     \label{fig:qualitative_nyud}
\end{figure*}

To confirm the superior performance of the proposed TaskExpert, we visualize its task-specific prediction maps on the testing set, and compare them with the output of the previous best state-of-the-art method~(\ie InvPT~\cite{invpt2022}) on the PASCAL-Context~\cite{everingham2010pascal} and NYUD-v2~\cite{silberman2012indoor} datasets. The results are presented in Figure~\ref{fig:qualitative_pascal2} and Figure~\ref{fig:qualitative_nyud}, respectively. The qualitative study demonstrates that our TaskExpert can generate finer results, particularly in semantic segmentation and human parsing.

\clearpage

{\small
\bibliographystyle{unsrt}
\bibliography{refers}
}

\end{document}